\pdfoutput=1
\documentclass[a4paper,10pt]{article}

\usepackage{fullpage} 
\usepackage{parskip} 
\usepackage{tikz} 
\usepackage{amsmath,tabularx}
\usepackage{hyperref}
\usepackage{graphicx}
\usepackage[font=footnotesize]{caption}
\usepackage{array}
\usepackage{multicol}
\usepackage{amssymb}
\usepackage[ruled,vlined,linesnumbered]{algorithm2e}
\usepackage{mathrsfs}
\usepackage{bm}
\SetAlFnt{\footnotesize}
\usepackage{adjustbox}
\graphicspath{ {images/} }
\usepackage{color}
\usepackage{hyperref}
\usepackage{booktabs}
\usepackage{multirow}
\usepackage[numbers]{natbib}
\usepackage{numprint}
\usepackage{tikz}
\usetikzlibrary{bayesnet}
\usetikzlibrary{arrows}
\usepackage{subcaption}
\usepackage{subfiles}
\usepackage{mathtools}
\usepackage{breqn}
\usepackage{lscape}

\newcommand{\argmin}{\operatornamewithlimits{arg\ min}}

\begin{document}

\title{Learning Latent Representations of\\ Bank Customers With \\ The Variational Autoencoder}
\author{Rogelio A. Mancisidor$^{a,b,\ast}$, Michael Kampffmeyer $^a$, 
Kjersti Aas $^c$, Robert Jenssen $^a$\\
\\
\normalsize{$^{a}$UiT Machine Learning Group, Faculty of Science and Technology,}\\
\normalsize{Department of Physics and Technology, University of Troms{\o}, 9019 Troms{\o} Norway,}\\
\normalsize{$^b$Credit Risk Models, Santander Consumer Bank AS, 1325 Lysaker Norway}\\
\normalsize{$^{c}$Statistical Analysis, Machine Learning and Image Analysis,}\\
\normalsize{Norwegian Computing Center, 0373 Oslo Norway}\\
\\
\normalsize{$^\ast$Corresponding author; E-mail: rogelio.a.mancisidor@uit.no}
\
}

\maketitle

\renewenvironment{abstract}
{\begin{quote}
\noindent \rule{\linewidth}{.5pt}\par{\bfseries \abstractname.}}
{\noindent 
\rule{\linewidth}{.5pt}
\end{quote}
}

\begin{abstract}
Learning data representations that reflect the customers' creditworthiness can improve marketing campaigns, customer relationship management, data and process management or the credit risk assessment in retail banks. In this research, we adopt the Variational Autoencoder (VAE), which has the ability to learn latent representations that contain useful information. We show that it is possible to steer the latent representations in the latent space of the VAE using the Weight of Evidence and forming a specific grouping of the data that reflects the customers' creditworthiness. Our proposed method learns a latent representation of the data, which shows a well-defied clustering structure capturing the customers' creditworthiness. These clusters are well suited for the aforementioned banks' activities. Further, our methodology generalizes to new customers, captures high-dimensional and complex financial data, and scales to large data sets. 


\end{abstract}

\section{Introduction}\label{intro}
Banks need to estimate the creditworthiness of both customers and applicants to improve marketing campaigns, customer relationship management, data and process management or the credit risk assessment \cite{anderson2007credit}. Customers with high creditworthiness can obtain higher credit limits or get offered new financial products, while applicants with low creditworthiness may not be eligible for credit at all. Therefore, learning data representations that can support these activities is critical. 

The Variational Autoencoder (VAE) \cite{kingma2013auto,rezende2014stochastic} has shown promising results in different research domains. The powerful information embedded in its latent space has been documented e.g., in health analytics \cite{rampasek2017dr,bioinformatics18,way2017evaluating,way2017extracting}, in speech emotion recognition (SER) \cite{latif2017variational}, and in natural language processing (NLP) \cite{bowman2015generating,su2018neural}, among others. Additionally, research has been conducted where the VAE has been modified to improve its feature learning properties, e.g. \cite{bouchacourt2018multi,higgins2017beta,hsu2017learning,su2018neural}. However, to the best of our knowledge there is no previous work on data representations of bank customers using the VAE.

Inspired by the previous results in other research fields and the lack of research on learning data representations of bank customers, we adopt the VAE and the Auto Encoding Variational Bayesian (AEVB) algorithm \cite{kingma2013auto} to learn a data representation that is useful to support the banking activities. Our proposed method is able to steer the latent embeddings in the VAE by transforming the input data into a meaningful representation, and by creating a specific grouping of the data. Hence, the focus of this research is on the effective manifold learning capabilities of the VAE \cite{Goodfellow-et-al-2016}, which captures valuable information in the latent space.

The main advantage of our method is that it provides a data representation in the low-dimensional latent space of the VAE, which can be visualized and it suggests well-defined clustering structures. These clusters are well suited to the bank industry  given that they encapsulate customers' creditworthiness. In addition, customers' creditworthiness is perfectly ranked across the dimensions of the latent space. Further, using the generative properties of the VAE, we can draw the latent space of unseen customers and map them into an existing cluster without the need of further supervision. Finally, this technique is able to analyse high-dimensional financial data with complex non-linear relationships \cite{khandani2010consumer}, and it scales to large data sets \cite{kingma2013auto}. 

This paper is organized as follows. Section \ref{sec_literature} reviews the related work where the VAE has been used to learn data representations in different research fields, while Section \ref{sec_ae} introduces variational inference and the VAE. In Section \ref{sec_learning} we explain the data transformation used to learn latent representations of bank customers and Section \ref{sec_results} presents our experiments and findings. Finally, Section \ref{sec_conclusion} presents the main conclusions in this paper.

\section{Related Work}\label{sec_literature}
Methods to learn data representations from the input data can be divided into probabilistic graphical models (PGMs) and neural network-based models \cite{bengio2013representation}. This data representation plays an important role in the results we can achieve in detection or classification tasks \cite{bengio2013representation, lecun2015deep,zhong2016overview}. The ability to express general-purpose priors, such as natural clustering or spatial coherence, among others, is what make data representations to be good \cite{bengio2013representation}.

Further, PGMs aim to learn latent representations $\bm{z}$, which are able to describe the input data $\bm{x}$. This is done by modelling their joint distribution $p(\bm{z},\bm{x})$. Depending on how this joint distribution is constructed, PGMs can be divided in directed or undirected graphical models \cite{bengio2013representation}. 

The Variational Autoencoder (VAE) \cite{kingma2013auto,rezende2014stochastic} is an influential (unsupervised) directed probabilistic graphical model, which has been widely used to learn meaningful latent representations of the input data. For example, latent representations of gene expression data are used in  \cite{way2017evaluating} for cancer prediction. The results show that the VAE latent features are useful to predict cancer and its predictive power is similar to other data transformation methods, e.g. principal component analysis (PCA) \cite{pearson1901liii}.

Latent representations in the VAE have also been used for predictions in a semi-supervised context. In \cite{rampasek2017dr}, latent representations for pre-treatment and post-treatment gene expression are use to predict drug response. Their proposed model achieves higher performance relative to Ridge logistic regression \cite{hoerl1970ridge} using the original input data. In addition, PCA transformations are used in three different classifiers to predict drug responses, but their performance, in most of the experiments, is not better relative to Ridge regression and the VAE model.

Classification of speech emotion is another example where latent representations of the input data have been successfully used for classification. Using Long Short Term Memory (LSTM) networks to classify emotion,  \cite{latif2017variational} compares the predictive power of data transformations using VAE and regular Auto Encoders. Speech emotion prediction is more accurate when the latent representations in the VAE are used as predictors. The classification results are further improved by using latent representations obtained with conditional VAE \cite{sohn2015learning}.

In another classification study, \cite{bioinformatics18} train logistic regression models, on t-SNE \cite{hinton2003stochastic} embeddings of high-dimensional VAE latent variables, to classify tumours. Their results show that the latent embeddings in the VAE learn a biological relevant information and successfully classify disease sub-types. Both works in \cite{latif2017variational,bioinformatics18} build upon the \textit{Tybalt} model \cite{way2017extracting}. The Tybalt exploits the data transformation capabilities of the VAE to generate latent representations of gene expression data. 

The VAE has also been used in the natural language processing field. Studying bilingual word embeddings,  \cite{su2018neural} use the VAE to generate latent representations, which explicitly induce the underlying semantics of bilingual text. Their model is able to learn a hidden representation of paired bag-of-words sentences. Furthermore, in \cite{bowman2015generating} recurrent neural networks are combined with the VAE to model text data. The latent transformations are able to generate coherent sentences. In addition, the proposed model in this research is able to impute missing words in text corpus. 

Research has also been conducted on modifying the original VAE aiming to improve the quality of the learned latent representations. In \cite{higgins2017beta}, for example, the authors add an hyperparamter $\beta$ to the VAE, which limits the capacity of the latent information channel and impose an emphasis on learning statistically independent latent factors. Hence, the model is able to learn disentangled factors of variation. 

In \cite{bouchacourt2018multi} the concept of supervision in VAE is introduced. The authors group the input data, aiming to learn representations of the data that reflect the semantics behind a specific grouping of the data. In other words, the grouping makes it possible to learn a semantically useful data transformation. Similarly, \cite{hsu2017learning,su2018neural} use supervision but in the latent space. Both works \cite{hsu2017learning,su2018neural}, manipulate the latent representations arithmetically to decompose the latent representation into different attributes. 

In this research, as in \cite{bouchacourt2018multi,hsu2017learning,su2018neural}, we introduce a supervision stage in the VAE. In this stage, we form groups that share a common factor of variation. The difference in our method is that the grouping is derived from the class label, see Section \ref{sec_learning}. This means that our proposed method is a semi-supervised representation learning model where we indirectly steer the data transformation using a specific grouping of the input data. Finally, we only focus on learning a data representation of bank customers' data that is able to capture the customers' creditworthiness in the latent space of the VAE, and not in the predictive power of such representations. 

\section{The Variational Autoencoder}\label{sec_ae}
\subsection{Variational Inference}\label{sec_vi}
In the rest of the paper we use the following notation. We consider i.i.d. data $\{\bm{x}_i\}_{i=1}^n$ where $\bm{x}_i \in \mathbb{R}^{d_x}$ is the customers data, e.g. income, age, marital status etc. Further, the latent variables $\{\bm{z}_i\}_{i=1}^n$ where $\bm{z}_i \in \mathbb{R}^{d_z}$ are the data transformation of $\bm{x}_i$. The subscript $i$ is dropped whenever the context allows for it. 

The latent variable in the joint density $p(\bm{x},\bm{z})$ is drawn from a prior density $p(\bm{z})$ and then it is linked to the observed data through the likelihood $p(\bm{x}|\bm{z})$. Inference amounts to conditioning on data and computing the posterior $p(\bm{z}|\bm{x})$ \cite{vi_review}.

The problem is that the posterior distribution $p(\bm{z}|\bm{x})$ is intractable in most cases. Note that
\begin{equation}
p(\bm{z}|\bm{x}) = \frac{p(\bm{z},\bm{x})}{p(\bm{x})},
\label{true_post}
\end{equation}
involves the marginal distribution $p(\bm{x})=\int p(\bm{z},\bm{x}) d\bm{z}$. This integral, called the \textit{evidence}, in some cases requires exponential time to be evaluated since it considers all configurations of latent variables. In other instances, it is unavailable in a closed form \cite{vi_review}.

Variational Inference (VI) copes with this kind of problem by minimizing the Kullback-Leibler (KL) divergence\footnote{The KL divergence is a measure of the proximity between two densities, e.g. $KL[q(\cdot)||p(\cdot)]$, and it is commonly measured in bits. It is non-negative and it is minimized when $q(\cdot)=p(\cdot)$.} between the true posterior distribution $p(\bm{z}|\bm{x})$ and a parametric function $q(\bm{z}|\bm{x})$, which is chosen among a set of densities $\Im$ \cite{vi_review}. This set of densities is parameterized by \textit{variational parameters} and they should be flexible enough to capture a density close to $p(\bm{z}|\bm{x})$ and, in addition, be simple for efficient optimization. The parametric density which minimizes the KL divergence is 
\begin{equation}
q^*(\bm{z}|\bm{x})=\argmin_{q(\bm{z}|\bm{x})\in\Im} KL[q(\bm{z}|\bm{x})||p(\bm{z}|\bm{x})].
\label{KL_argmin}
\end{equation}
Unfortunately, Equation \ref{KL_argmin} cannot be optimized directly since it requires computing a function of $p(\bm{x})$. To see this, let us expand the KL divergence using the Bayes' theorem and noting that $p(\bm{x})$ does not depend on $\bm{z}$
\begin{align}
KL[q(\bm{z}|\bm{x})||p(\bm{z}|\bm{x})] =& E_{\bm{z} \sim q}[\log q(\bm{z}|\bm{x}) - \log p(\bm{z}|\bm{x})] \nonumber \\
=&  E_{\bm{z} \sim q}[\log q(\bm{z}|\bm{x}) - \log p(\bm{x},\bm{z})]+\log p(\bm{x}). 
\label{KL_intrac}
\end{align}
Given that Equation \ref{KL_intrac} cannot be optimized directly, VI optimizes the alternative objective function 
\begin{align}
E_{\bm{z} \sim q}[\log p(\bm{x},\bm{z}) - \log q(\bm{z}|\bm{x})]= & E_{\bm{z} \sim q}[\log p(\bm{z}) + \log p(\bm{x}|\bm{z}) - \log q(\bm{z}|\bm{x})] \nonumber \\
=& E_{\bm{z} \sim q}[\log p(\bm{x}|\bm{z})]-KL[q(\bm{z}|\bm{x})||p(\bm{z})] \nonumber \\
=& ELBO.
\label{ELBO}
\end{align}
From Equations \ref{KL_intrac} and \ref{ELBO} we have that
\begin{align}
\log p(\bm{x})  = KL[q(\bm{z}|\bm{x})||p(\bm{z}|\bm{x})] + ELBO.
\label{eq_log_loglikelohood}
\end{align}
Since the KL divergence is non-negative, the expression in Equation \ref{ELBO} is called \textit{the evidence lower bound} (ELBO). Noting that the ELBO is the negative KL divergence in Equation \ref{KL_intrac} plus the constant term $\log p(\bm{x})$, it follows that maximizing the ELBO leads to minimizing Equation \ref{KL_argmin}. 

It is worth mentioning that the term $KL[q(\bm{z}|\bm{x})||p(\bm{z})]$ encourages variational densities to be close to the prior distribution, while the term $E_{\bm{z} \sim q}[\log p(\bm{x}|\bm{z})]$ encourages densities that place their mass on configurations of the latent variables that explain the observed data. The interested reader is referred to \cite{vi_review,doersch2016tutorial} for further details. 

\begin{figure}[t!]
    \centering
	\includegraphics[scale=0.61]{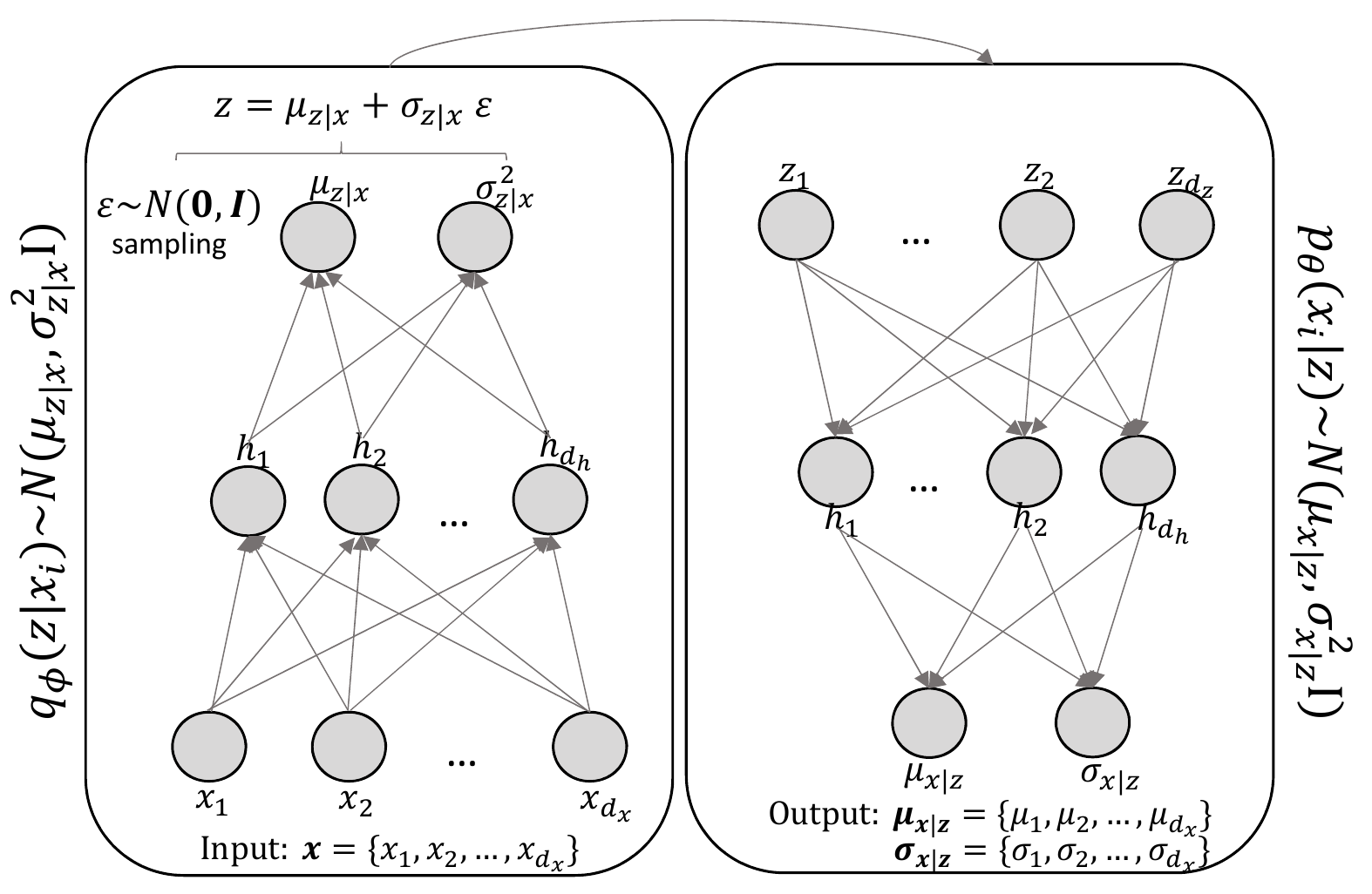}	
	\caption{Graphical representation of the VAE. The multilayer perceptron network to the left corresponds to the probabilistic encoder $q_{\phi}(\bm{z}|\bm{x})$ where $\bm{x} \in \mathbb{R}^{d_{\bm{x}}}$ is the network input. The output of the network are the parameters in \smash{$q_{\phi}(\bm{z}|\bm{x}) \sim \mathcal{N}(\bm{\mu}_{\bm{z}|\bm{x}}, \bm{\sigma}_{\bm{z}|\bm{x}}^2 \bm{I})$}. Note that $\epsilon \sim \mathcal{N}(0,\bm{I})$ is drawn outside the network in order to use gradient descent and backpropagation optimization techniques. Similarly, the feedforward network to the right corresponds to the probabilistic decoder $p_{\theta}(\bm{x}|\bm{z})$. In this case, the input are the latent variables $\bm{z} \in \mathbb{R}^{d_{\bm{z}}}$ and the network output are the parameters in \smash{$p_{\theta}(\bm{x}|\bm{z}) \sim \mathcal{N}(\bm{\mu}_{\bm{x}|\bm{z}}, \bm{\sigma}_{\bm{x}|\bm{z}}^2 \bm{I})$}. For readability purposes we do not specify the weights $\bm{\phi}$ and $\bm{\theta}$ in the decoder and encoder respectively. However, these parameters are represented by the lines joining the nodes in the networks plus a bias term attached to each node.}
	\label{vae_diagram}
\end{figure}

\subsection{The Variational Autoencoder and AEVB algorithm}
The Variational Autoencoder, see Figure \ref{vae_diagram}, is a generative model, which aims to learn the distribution of the input data $\bm{x}$. This means that the VAE can sample from a distribution that it is similar to the one that have generated $\bm{x}$. In addition, the VAE assumes that latent variables $p(\bm{z}) \sim \mathcal{N}(\mathbf{0},\mathbf{I})$ govern the distribution of $\bm{x}$. In this research, the input data $\bm{x}$ represents the customer data, or a specific grouping of it, and the data transformation of such data is generated by $q(\bm{z}|\bm{x})$. This data representation of the customer data should capture the customers' creditworthiness. In this section, we will show how the VAE approximates Equation \ref{eq_log_loglikelohood} by maximizing the ELBO. This is done using multilayer perceptron (MLP) networks and stochastic gradient optimization.

The MLPs, which optimze the ELBO, estimate the parameters $\bm{\mu}_{\cdot}$ and $\bm{\sigma}_{\cdot}^2$ in the density functions $p_{\bm{\theta}}(\bm{x}|\bm{z})$ and $q_{\bm{\phi}}(\bm{z}|\bm{x})$, i.e. $p_{\bm{\theta}}(\bm{x}|\bm{z}) \sim \mathcal{N}(\bm{\mu}_{\bm{x}|\bm{z}}^{},\bm{\sigma}_{\bm{x}|\bm{z}}^2\bm{I})$ and $q_{\bm{\phi}}(\bm{z}|\bm{x}) \sim \mathcal{N}(\bm{\mu}_{\bm{z}|\bm{x}}^{},\bm{\sigma}_{\bm{z}|\bm{x}}^2\bm{I})$. Note that given that the output of the MLPs are $\bm{\mu}_{\cdot}$ and $\bm{\sigma}_{\cdot}^2$, the stochastic gradient optimization is on $\bm{\theta}$ and $\bm{\phi}$, which are the weights in the MLPs. By doing this, the VAE learns the values of $\bm{\mu}_{\cdot}$ and $\bm{\sigma}_{\cdot}^2$ that maximize the ELBO\footnote{It is possible to specify other distributions for $p(\cdot)$ and $q(\cdot)$. However, Gaussian distributions are appropriate for our data sets, and we assume a diagonal covariance matrix as in the original VAE.}.

Specifically, assuming the set of i.i.d vectors $\{\bm{x}_i,...,\bm{x}_n\}$, the Auto Encoding Variational Bayesian (AEVB) algorithm \cite{kingma2013auto} learns the parameters $\bm{\theta},\bm{\phi}$ jointly using MLP networks, and by performing stochastic gradient descent on the 
\begin{equation}
    ELBO(\bm{\theta},\bm{\phi},\bm{x}_i) =  E_{\bm{z} \sim q}[\log p_{\bm{\theta}}(\bm{x}_i|\bm{z})]-KL[q_{\bm{\phi}}(\bm{z}|\bm{x}_i)||p(\bm{z})]
    \label{alt_elbo}
\end{equation}
for the \textit{i}'th customer. Therefore, the MLPs for $q_{\bm{\phi}}(\bm{z}|\bm{x})$ and $p_{\bm{\theta}}(\bm{x}|\bm{z})$ in Figure \ref{vae_diagram} have the following form
\noindent\begin{minipage}{.5\linewidth}
\begin{align}
\bm{h} =& \tanh(\bm{W}_1 \bm{x}_i + \bm{b}_1), \nonumber \\
\bm{\mu}_{\bm{z}|\bm{x}} =& \bm{W}_2 \bm{h} + \bm{b}_2, \nonumber  \\
\log \bm{\sigma}_{\bm{z}|\bm{x}}^2 =& \bm{W}_3 \bm{h} + \bm{b}_3, \nonumber  \\
\bm{z}_i =& \bm{\mu}_{\bm{z}|\bm{x}} + \bm{\sigma}_{\bm{z}|\bm{x}} \odot \bm{\epsilon} \nonumber 
\end{align}
\end{minipage}%
\begin{minipage}{.5\linewidth}
\begin{align}
\bm{h} =& \tanh(\bm{W}_4 \bm{z}_i + \bm{b}_4), \nonumber \\
\bm{\mu}_{\bm{x}|\bm{z}} =& \bm{W}_5 \bm{h} + \bm{b}_5, \nonumber  \\
\log \bm{\sigma}_{\bm{x}|\bm{z}}^2 =& \bm{W}_6 \bm{h} + \bm{b}_6, \nonumber  \\
\hat{\bm{x}}_i =& \bm{\mu}_{\bm{x}|\bm{z}} + \bm{\sigma}_{\bm{x}|\bm{z}} \odot \bm{\epsilon} , 
\label{eq_networs}
\end{align}
\end{minipage}

where $\bm{\epsilon} \sim \mathcal{N(\mathbf{0},\mathbf{I})}$, $\odot$ is the element-wise product, $\bm{\phi} =\{\bm{W}_1,\bm{W}_2,\bm{W}_3,\bm{b}_1,\bm{b}_2,\bm{b}_3\}$ and $\bm{\theta} =\{\bm{W}_4,\bm{W}_5,$ $\bm{W}_6,\bm{b}_4,\bm{b}_5,\bm{b}_6\}$ are the unknown parameters in the MLPs for $q_{\bm{\phi}}(\bm{z}|\bm{x})$ and $p_{\bm{\theta}}(\bm{x}|\bm{z})$ respectively.

It is worth mentioning that the latent variable $\bm{z}$ has been reparametrized as a deterministic and differentiable system. The reason is that we need to backpropagate the term $E_{\bm{z} \sim q}[\log p_{\bm{\theta}}(\bm{x}_i|\bm{z})]$ in Equation \ref{alt_elbo}. Without the reparametrization, $\bm{z}$ would be inside a sampling operation which cannot be propagated. This means that the AEVB algorithm actually takes the gradient of $E_{\bm{\epsilon} \sim \mathcal{N(\mathbf{0},\mathbf{I})}}[\log p_{\bm{\theta}}(\bm{x}_i|\bm{z}_i= \bm{\mu}_i + \bm{\sigma}_i \odot \bm{\epsilon} )]$. The proof of this result can be found in \cite{kingma2013auto}. 

Note that $q_{\bm{\phi}}(\bm{z}|\bm{x})$ generates latent variables given $\bm{x}$ and $p_{\bm{\theta}}(\bm{x}|\bm{z})$ converts them into its original representation. Hence, the former is referred as probabilistic encoder and the latter as probabilistic decoder.

\section{Learning Latent Representations}\label{sec_learning}
In this section we introduce the motivation for the specific grouping of data that we use to steer a data representation, which encapsulates the customers' creditworthiness in the latent space of the VAE. The presumption is that given that the AEVB algorithm has converged to the optimal variational density $q^*(\bm{z})$, the latent space should have learned a data representation, which encapsulates the customers' creditworthiness. Otherwise, the reconstruction would have failed, and the algorithm would not have converged to $q^*(\bm{z})$ in the first place. 

To quantify creditworthiness, let us first define the ground truth class
\begin{equation}
y=
    \left\{
    \begin{array}{ll}
      1 & \text{if at least 90 days past due}  \\
      0 & \text{otherwise}.
    \end{array} 
    \right. \
    \label{default_flag}
\end{equation}
At least 90 days past due, or just 90+dpd, refers to the customers' payment status, which is known after the performance period is over \footnote{The performance period is the time interval in which if customers are at any moment 90+dpd, then their ground truth class is $y=1$. Frequently, 12 and 24 months are time intervals used for the performance period. Further, the performance period starts at the moment an applicant signs the loan contract.}. This definition is aligned with the Basel II regulatory framework \cite{anderson2007credit}. 

Let $C_j = \{c_{j,1},c_{j,2},...,c_{j,n_j}\}$ be the \textit{j'th} set of customers with class labels  $Y_j = \{y_{j,1},y_{j,2},...,y_{j,n_j}\}$. Hence, 
\begin{equation}
    dr_{C_j} = \frac{\sum_i^{n_j} [y_{j,i}=1]}{n_{j}},
    \label{eq_diffdr}
\end{equation}
where $[\cdot]$ is the Iverson bracket, is the default rate of $\textit{j}$'th group of customers.

Given that $dr_{C_j}>dr_{C_l}$, we say that group $C_j$ has lower creditworthiness compared to group ${C_l}$. In other words, customers in $C_j$ have, on average, higher probability of default. Therefore, in order to identify $L$ segments with a different propensity to fall into financial distress, we need to find segments where the average probability of default is different from the rest of the groups. Mathematically, we want to learn a data representation that satisfies 
\begin{equation}
    dr_{C_j} \neq dr_{C_l}, \quad for \quad j,l = 1,2,...,L \ and \ j\neq l.
    \label{eq_diffdp}
\end{equation}
Now it should be clear that the data transformation $f(\cdot)$ that we are looking for, needs to incorporate the class label $y$. In this way, the latent space in the VAE should generate codes that also contain information about $y$. Otherwise, those codes will fail to reproduce $f(\bm{x}|y)$.  

\begin{table}[t!]
\centering 
\caption{Weight of Evidence transformation of the variable age. The top panel shows the fine classing approach, while the bottom panel shows the coarse approach where only three groups are created.}
\begin{adjustbox}{width=\textwidth}
\begin{tabular}{ccccccccc}
\hline
\multicolumn{9}{c}{Fine classing approach}\\
\hline
 Age & Count & Total Distribution & Goods & Distribution Goods  &  Bads & Distribution Bads & Bad Rate & WoE \\
\hline
Missing & 1 000 & 2.5\% & 860    & 2.38\%   & 140   & 3.65\%    & 14.00 \%  & -0.4272 \\
18-22 & 4 000 & 10\%    & 3 040  & 8.41\%   & 960   & 25.00\%   & 24.00 \%  & -1.0898 \\
23-26 & 6 000 & 15\%    & 4 920  & 13.61\%  & 1 080 & 28.13\%   & 18.00 \%  & -0.7261 \\
27-29 & 9 000 & 22.5\%  & 8 100  & 22.40\%  & 900   & 23.44\%   & 10.00 \%  & -0.0453 \\
30-35 & 10 000 & 25.0\% & 9 500  & 26.27\%  & 500   & 13.02\%   & 5.00 \%   & 0.7019 \\
36-44 & 7 000 & 17.5\%  & 6 800  & 18.81\%  & 200   & 5.21\%    & 2.86 \%   & 1.2839 \\
44+   & 3 000 & 7.5\%   & 2 940  & 8.13\%   & 60    & 1.56\%    & 2.00 \%   & 1.6493 \\
Total & 40 000 & 100\%  & 36 160 & 100\%    & 3 840 & 100\%     & 9.60 \%    & \\
\hline
\multicolumn{9}{c}{Coarse classing approach}\\
\hline
Missing &	1 000	& 2.5 \%	& 860	 &  2.38 \%	& 140	& 3.65 \%   &  14.00\%	&  -0.4272 \\
18-29	& 19 000	& 47.5 \%	& 16 060 &	44.41\%	& 2 940	& 76.56	\%  &  15.47 \% &  -0.5445 \\
30-44+	& 20 000	& 50\%	    & 19 240 &	53.20\%	& 760	& 19.79 \%  &  3.80	\%  &   0.9889 \\
Total	& 40 000	& 100\%	    & 36160	 &  100 \%  & 3840	& 100 \%    &  9.60\%	&    \\
\hline
\end{tabular}
\end{adjustbox}
\label{tbl_woe}
\end{table}

One such transformation is the Weight of Evidence\footnote{Originally, the WoE was introduced by Irving John Good in 1950 in his book \textit{Probability and the Weighing of Evidence} and it has been used in the logistic regression and Na\"ive Bayes literature, among others.} (WoE) \cite{anderson2007credit,siddiqi2012credit}, which is defined as
\begin{align}
    \log \frac{Pr(\bm{x}|y=0)}{Pr(\bm{x}|y=1)}.
    \label{woe_eq}
\end{align}

\subsection{The Weight of Evidence}
The WoE transformation has been used in credit scoring for a long time \cite{abdou2009genetic}, and it has become the standard in credit scoring models. The way to estimate it, given that the \textit{m}'th feature $x_m$ is continuous, is by dividing its values into $K$ bins $B_1, B_2, ..., B_K$. In the case of categorical variables, the different categories are already these bins. Hence, the WoE for the \textit{k}'th bin of the \textit{m}'th feature is
\begin{equation}
    WoE_{k,m} = \log \frac{Pr(x_m \in B_k|y=0)}{Pr(x_m \in B_k|y=1)} = \log \frac{\frac{1}{n}\sum_{i=1}^{n} [x_{i,m} \in B_{k,m} \ \text{and} \ y_i = 0]}{\frac{1}{n}\sum_{i=1}^{n} [x_{i,m} \in B_{k,m} \ \text{and} \ y_i = 1]},
\end{equation}
where $n$ is the total number of observations. Note that the number of bins can vary for different features. See chapter 16.2 in \cite{anderson2007credit} or chapter 6 in \cite{siddiqi2012credit} for more details. Table \ref{tbl_woe} shows the difference between fine and coarse classing. In the fine classing approach, we create $K$ bins, which provide the finest granularity. Then, fine bins with similar risk are binned into smaller groups resulting in the coarse classing. See  \cite{anderson2007credit} for more details.

We use the coarse classing WoE transformation\footnote{We will simply called it as WoE in the remaining of the paper for brevity.} of the input data $\bm{x}$ to tilt the latent space in the VAE towards configurations which encapsulate the propensity to fall into financial distress. 

\begin{figure}[t!]
    \centering
	\includegraphics[scale=0.65]{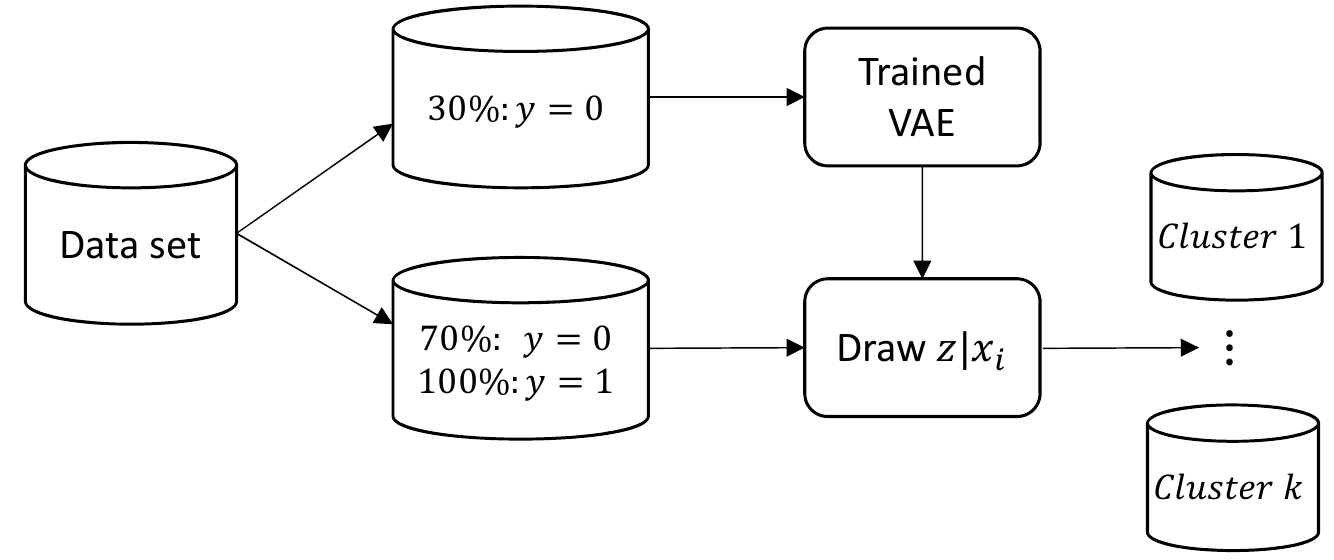}	
	\caption{Development methodology: We use 30\% of the majority class data for training the VAE. Once it is trained, we generate the latent variables for the remaining data.}
	\label{data_partition}
\end{figure}
\section{Experiments and Results}\label{sec_results}
Our goal with the experiments is threefold. First, we want to show in Section \ref{sec_data_rep} that our proposed method is able to learn \textit{better} representations compared to other methods and that these representations are able to encapsulate the customers' creditworthiness. Second, we introduce a specific grouping of the data in Section \ref{sec_learning}, which is used in the supervision stage. Hence, we want to show in Section \ref{sec_grouping} that not any other grouping is able to generate \textit{good} representations in the latent space. Third, we want to evaluate qualitatively in Section \ref{sec_interp} the data representation of bank customers analysing the salient dimensions in each clusters. 

\subsection{Data description}\label{data_sum}
We use three data sets in our experiments; a Norwegian and a Finnish car loan data set provided by Santander Consumer Bank Nordics and the public data set used in the Kaggle competition \textit{Give me some credit}\footnote{Website \url{https://www.kaggle.com/c/GiveMeSomeCredit}}. These data sets show applicants' status, financial and demographic factors at the time of application as well as the class label. The performance period for the real data sets is 12 months, while for the public data set it is 24 months. More details about the data sets can be found in Tables \ref{tbl_summary_dta}, \ref{tbl_variables_no} and \ref{tbl_variables}.

\subsection{Training the VAE and Generating Latent Representations}\label{sec_training_vae}
We train the VAE using the WoE as the input data, and using only the majority class ($y=0$) data. The reason is because we want to have a robust estimate for the default rate in the data representation that we are learning. In addition, using observations from the minority class ($y=1$) did not change the data representation in the latent space in our experiments, which is probably explained by the strong class imbalance in the three data sets. 

Hence, we use 30\% of the majority class to train the VAE. During training, we generate the latent space for the remaining 70\% of the majority class and 100\% of the minority class data, see Figure \ref{data_partition}. Based on the optimization of the ELBO, together with a heuristic visual comparison of the latent space in the training and test data sets, we select the optimal network architecture as well as the stopping criteria. It is worth mentioning that we observe that the shapes and proportions of the clusters in the training data are similar to the ones in the test data. This is a good indicator that our proposed model learns data representations that generalizes to unseen customers.

The VAE architectures that we tested are shown in Table \ref{all_archs}, and the final architecture IDs that we use are arch4, arch4 and arch1 for the Norwegian, Finnish and Kaggle data sets respectively. In addition, we use tanh activations in all hidden layers, linear and sigmoid activations in the $\mu$ output layer for the encoder and decoder respectively, and linear activations in all log $\sigma^2$ layers\footnote{We need to use different activation functions depending on the kind of variable that the MLP is handling. See chapter 6 in \cite{Goodfellow-et-al-2016} for more details.}. The MLP models are trained with the adagrad optimizer \cite{duchi2011adaptive} using constant 0.01 learning rate and 0.001 momentum. 

Finally, we use the expectation over the latent space for the \textit{i}'th customer 
\begin{equation}
    E[\bm{z}|\bm{x}_i]=\int_{-\infty}^{\infty}\bm{z} q_{\phi}(\bm{z}|\bm{x}_i)d\bm{z} = \bm{\mu}_{\bm{z}|\bm{x}}
    \label{eq_expect_z}
\end{equation}
as the data representation of bank customers in the latent space of the VAE. Note that it is simply the output in the encoder MLP network, see Equation \ref{eq_networs}. We also tried the Monte Carlo version of Equation \ref{eq_expect_z} using 100 samples of $\bm{z}_i$ and Equation \ref{eq_networs}. The results do not change. 

\begin{figure}[t!]
    \centering
	\includegraphics[scale=0.5]{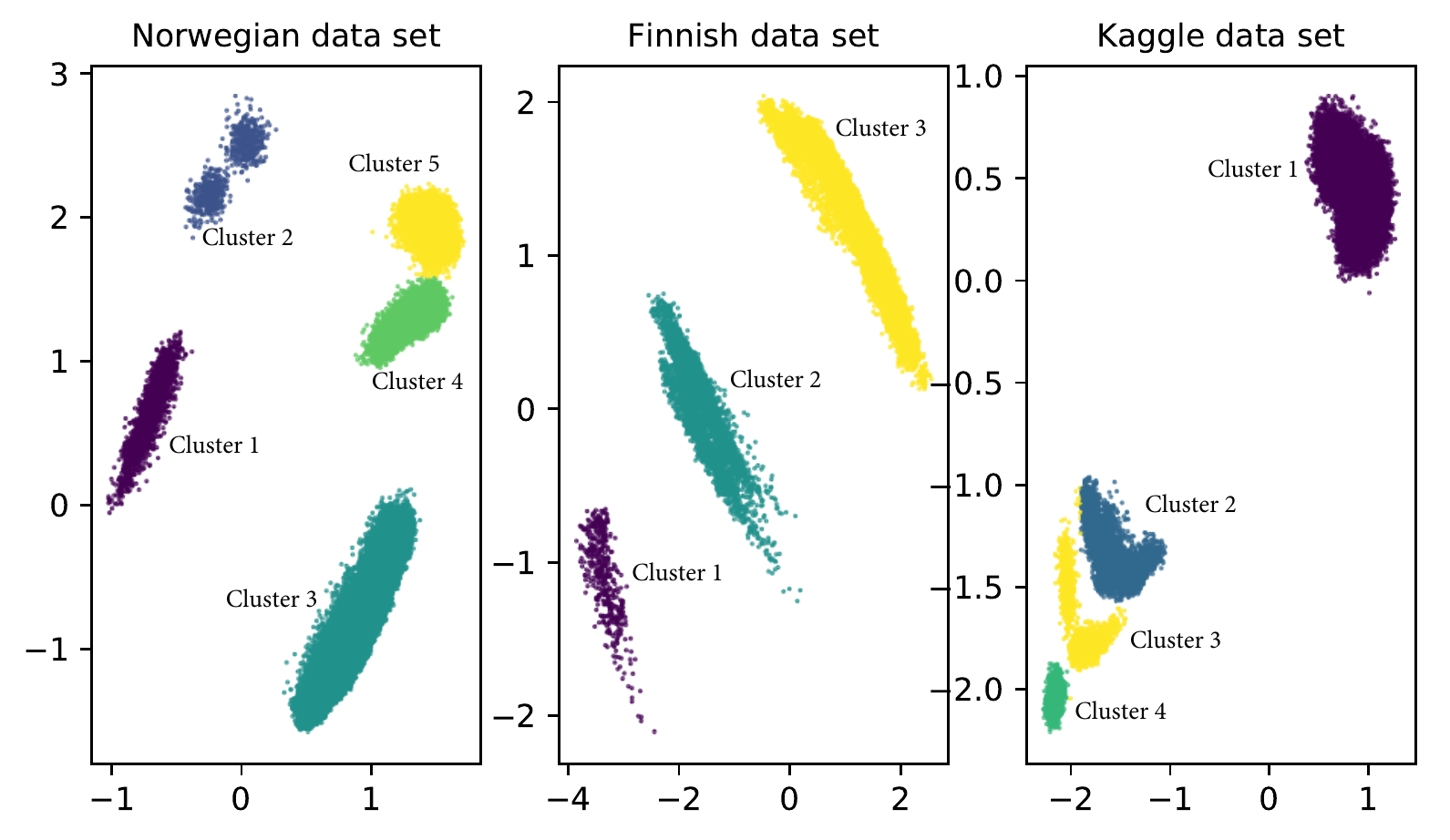}	
	\caption{Latent representation of bank customers.}
	\label{fig_latentspace}
\end{figure}

To further analyze the learned data representation of our method, we assign labels to the structure in the two-dimensional latent space. This task can be done manually using a set of \textit{if/else} rules given the well-defined clustering structure in the latent space. However, we propose an automated version, which is presented in Algorithm 1. The idea is to use the hierarchical clustering algorithm iteratively, generating only two clusters in each iteration. Always preserving the clustering structure in the learned data representation. For this purpose, Algorithm 1 specifies the minimum number of observations in each cluster, denoted by $n_{min}$. Similarly, the minimum Euclidean distance between the centroids in the two clusters needs to be specified, and it is denoted by $\rho$. These two paremeters are data dependent and should be selected in such a way that Algorithm 1 assigns cluster labels preserving the clustering structure in the latent space. The advantage of this approach is that the labelling happens automatically while we train the VAE. Finally, the results of our proposed data representations are shown in Figure \ref{fig_latentspace} and Table \ref{tbl_k_results_k_fi}.

\begin{algorithm}[t!]
\SetAlgoLined
\SetKwInOut{Input}{Input}
\SetKwInOut{Output}{Output}
\Input{$\bm{z}$,$n_{min}$,$\rho$}
\Output{cluster labels}
pending\_data = \{$\bm{z}$\} \;
 labels = ones(length($\bm{z}$)) \;
 \While  {EOF(pending\_data)==FALSE} { 
      \For {item in pending\_data} {
      labels = HierarchicalAlgorithm(pending\_data[item], k = 2) \;
      get centroids c1 and c2 \;
      split pending\_data[item] into C1 and C2 using labels  \;
      \If {$n1 > n_{min}$ AND $n2 > n_{min}$ AND  $||c1 - c2|| > \rho$}
      {
      update labels \;
      pending\_data.append = \{C1,C2\}
      }
      }
  }
  return labels
 \caption{Labelling the latent data representation of bank customers.}
 \label{alg_1}
\end{algorithm}

\subsection{Latent Data Representations}\label{sec_data_rep}
The first important result to highlight is that using the WoE transformation we learned a data representation with  well-defined clusters in the latent space for all three data sets. Analyzing the Norwegian car loan data, we see that about 82\% of all customers are in cluster 3, which is the cluster with the smallest default rate. This makes sense since the data set contains only 2 557 customers from the minority class. On the other hand, about 18\% of the customers are in clusters with relatively high default rate. Finally, we check whether the default rates are significantly different using the 99\% confidence interval for a binomial variable, i.e. $\hat{p} \pm 2.57 \sqrt{\hat{p}(1-\hat{p})/n}$, where $\hat{p}$ is the default rate in each cluster, $2.57$ is the corresponding critical value and $n$ is the total number of observations in the cluster. With the exception of clusters 1 and 5, the default rate for the other clusters are statistically different. See Table \ref{tbl_k_results_k_fi}.

For the Finnish and Kaggle data sets we observe the same pattern. The majority of the customers are in the cluster with the smallest default rate. However, about 10\% of the customers in the Kaggle data are in three clusters with very high default rates. Note that all default rates in the Kaggle data are significantly different. On the other hand, the confidence intervals for the default rates in cluster 1 and 2 for the Finnish data set overlap each other. This is driven by the relatively small number of defaults in the cluster 1, which increases the variance of their default rate estimate. 

\begin{landscape}
\begin{figure}
    \centering
    \includegraphics[scale=0.48]{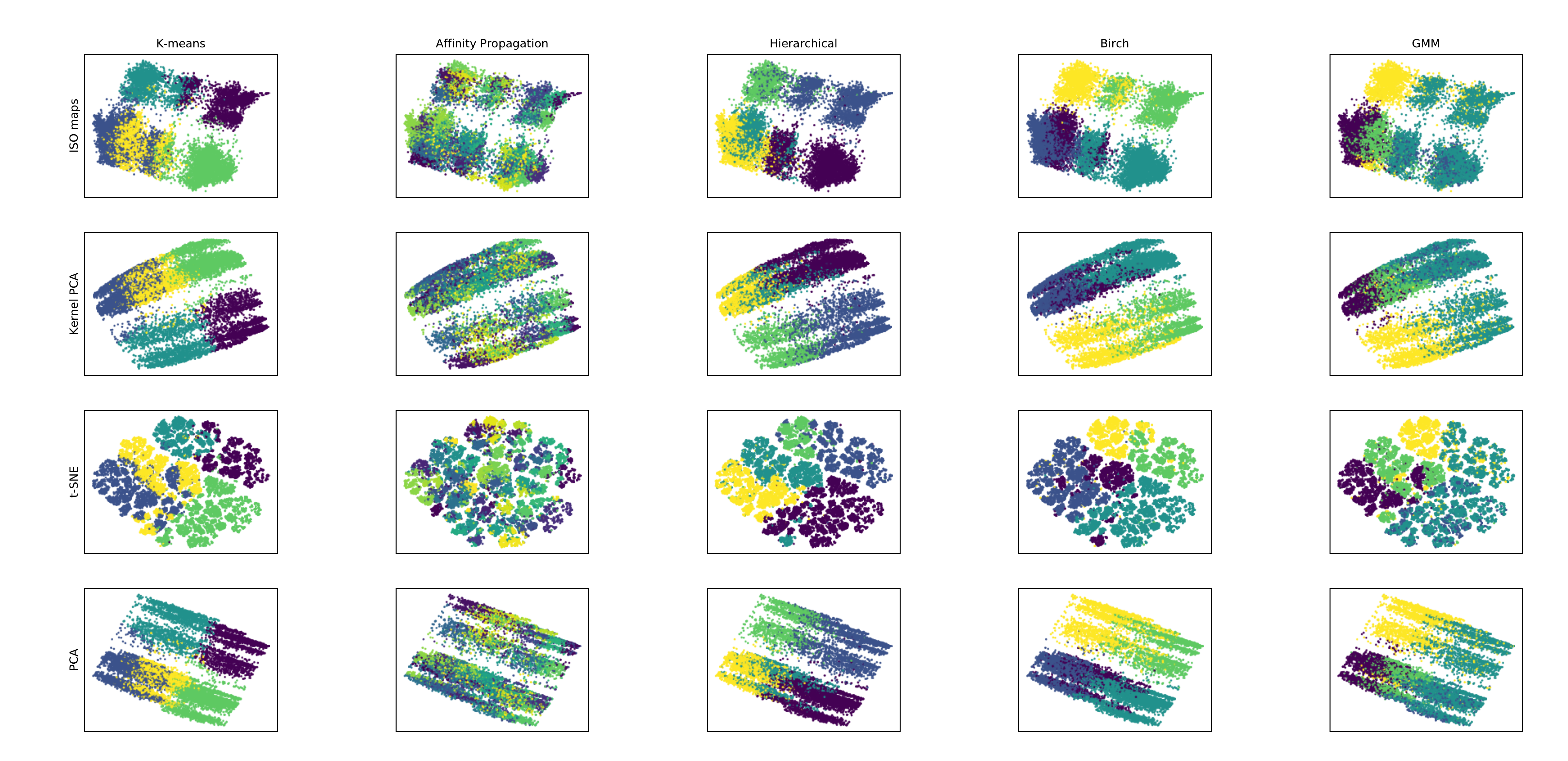}
    \caption{We use the k-means, affinity propagation, hierarchical, birch, and GMM algorithms to cluster the WoE transformations for the Norwegian data, specifying five clusters. Then, we reduce the original dimensional space for the WoE to two dimensions using isomaps, kernel PCA, t-SNE and PCA. Cluster labels are given by the colors.}
    \label{fig_transformations}
\end{figure}
\end{landscape}

Further, we use the k-means \cite{lloyd1982least}, affinity propagation \cite{frey2007clustering}, hierarchical \cite{ward1963hierarchical}, birch \cite{zhang1996birch} and GMM algorithms to cluster the WoE transformations for the Norwegian data set. We specify five clusters as suggested by the VAE. After the clustering is done, we reduce the original dimensional space for the WoE to two dimensions with isomaps \cite{tenenbaum2000global}, kernel PCA \cite{scholkopf1998nonlinear}, t-SNE \cite{hinton2003stochastic} and PCA \cite{pearson1901liii}. Figure \ref{fig_transformations} shows the resulting clusters represented by different colors. As can be seen from the figure, none of the clustering algorithms are able to generate non-overlapping clusters.

\begin{figure}[t!]
    \centering
    \includegraphics[scale=0.5]{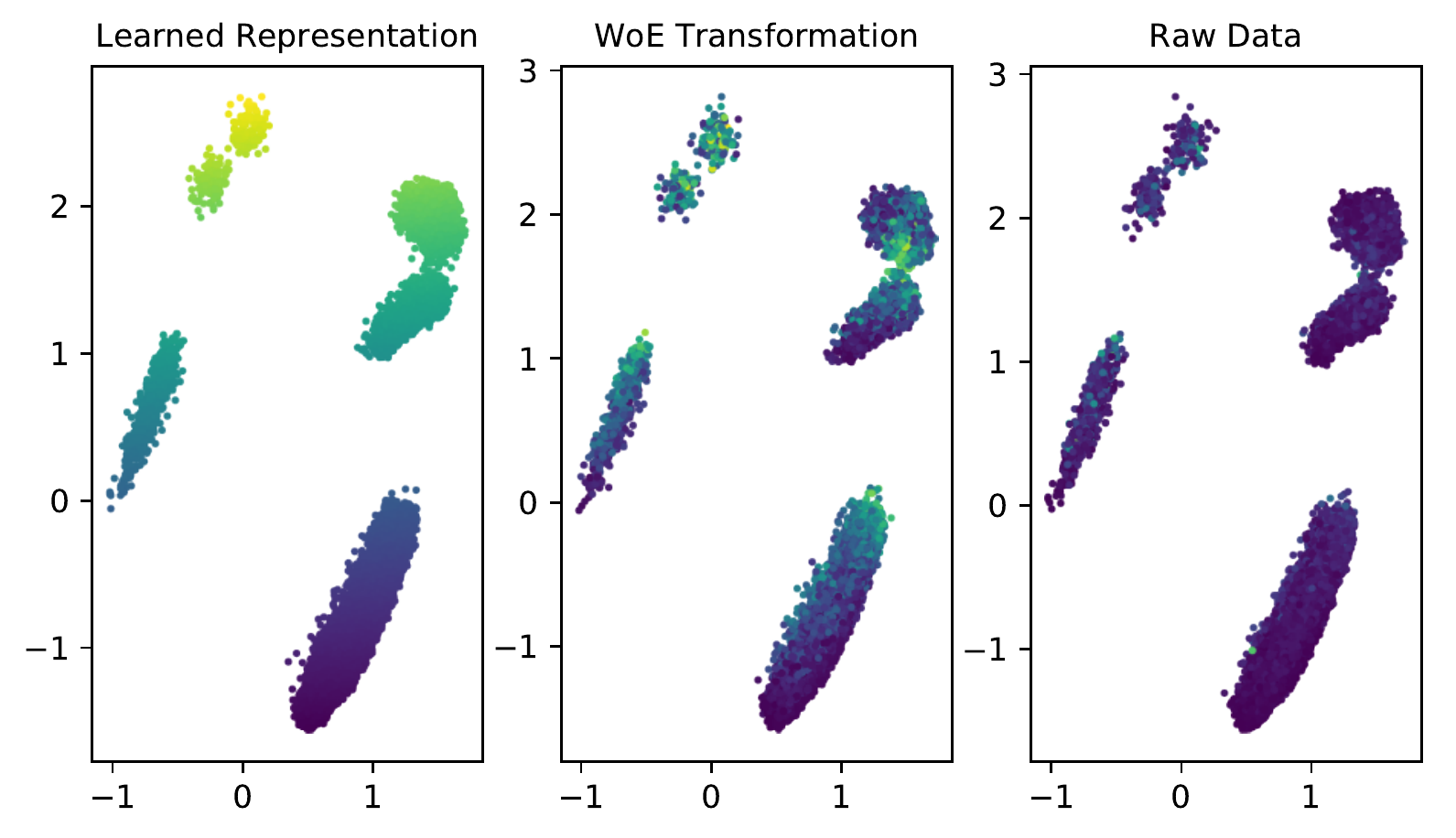}
    \caption{We estimate the default probability for 30\% of customers in the Norwegian data using the learned representation, WoE, and raw data. Further, we generate the latent space for these customers using the trained VAE. Finally, we use the three estimate values for default probability to create a colormap. Note that the left panel shows a smooth color transition.}
    \label{fig:my_label}
\end{figure}

Finally, we estimate the default probability for customers in the Norwegian data set using three different input data: i) the learned data representation of the VAE, ii) the WoE transformation, and iii) the raw data. We use 70\% of the data to estimate the default probability of the remaining 30\% of the data. Further, we use the trained VAE from Section \ref{sec_training_vae} to generate the latent space of the customers for whom we estimated the default probability. In Figure \ref{fig:my_label}, we show the learned data representation for these customers and we use the three estimated values for the default probability to create a colormap. It is interesting to see that the default probability estimated with the learned data representation reveals a smooth color transition. On the other hand, when the default probability is estimated with the WoE or with the raw data, the colormaps show a relatively random pattern. This result shows that our proposed method is not only able to learn a data representation of customers data, which shows a well-defined clustering structure and captures the customers' creditworthiness, but which also ranks the default probability across the two dimensions of the latent space. 

The well-defined clustering structure of the data representation in the latent space and its ability to capture customers' creditworthiness, allows our proposed method to generate good representations. These representations express general priors that are particularly useful in the bank industry.

\begin{table}[t!]
\centering 
\begin{adjustbox}{width=\textwidth}
\begin{tabular}{|c|cccc|cccc|cccc|}
\hline
 & \multicolumn{4}{c|}{\large{Norwegian car loan}} & \multicolumn{4}{c|}{\large{Finnish car loan }} & \multicolumn{4}{c|}{\large{Kaggle}} \\
\hline
Cluster & Default rate & 99\% CI & Nr. Customers & Nr. $y=1$ & Default rate & 99\% CI & Nr. Customers & Nr. $y=1$ & Default rate  & 99\% CI & Nr. Customers & Nr. $y=1$\\
\hline
1 & 5.30\%  &           & 2 206   & 117    &  5.93\%    &        & 438    & 26     & 5.47\%     & (***)  & 97 434  &  5 327 \\
2 & 11.24\% &  ***      & 774     & 87     & 3.76\%     & ***    & 6 067  & 228    & 33.13\%    & ***(***)  & 6 121   &  2 028   \\
3 &  1.39\% &  (***)    & 109 969 & 1539   & 0.91\%     & (***)  & 75 536 & 685    & 55.68\%    & ***(***)  & 2 026   &  1 128   \\
4 &  2.89\% &  ***(***) & 11 450  & 331    &            &        &        &        &  63.58\%   & ***  & 2 427   &  1 543   \\
5 & 5.30\%  &           & 9 106   & 483    &  &        &        &        &     &  &   &   \\
\hline
\end{tabular}
\end{adjustbox}
\caption{Default rates $dr_{C_j}$ for the different clusters $C_j$ in the data representation of bank customers. The 99\% confidence intevals are shown in the second column for each data set. Non-overlapping lower bounds are denoted outside parenthesis, while non-overlapping upper bounds are within parenthesis. Note that for the cluster with the lowest (highest) default rate we do not need to verify the lower (upper) bound.}
\label{tbl_k_results_k_fi}
\end{table}

Specifically, our proposed data representation is able to learn the natural clustering and spatial coherence of creditworthiness in the customers data. 

\subsection{Grouping of the Input Data}\label{sec_grouping}
\begin{figure}[t!]
    \centering
	\includegraphics[scale=0.5]{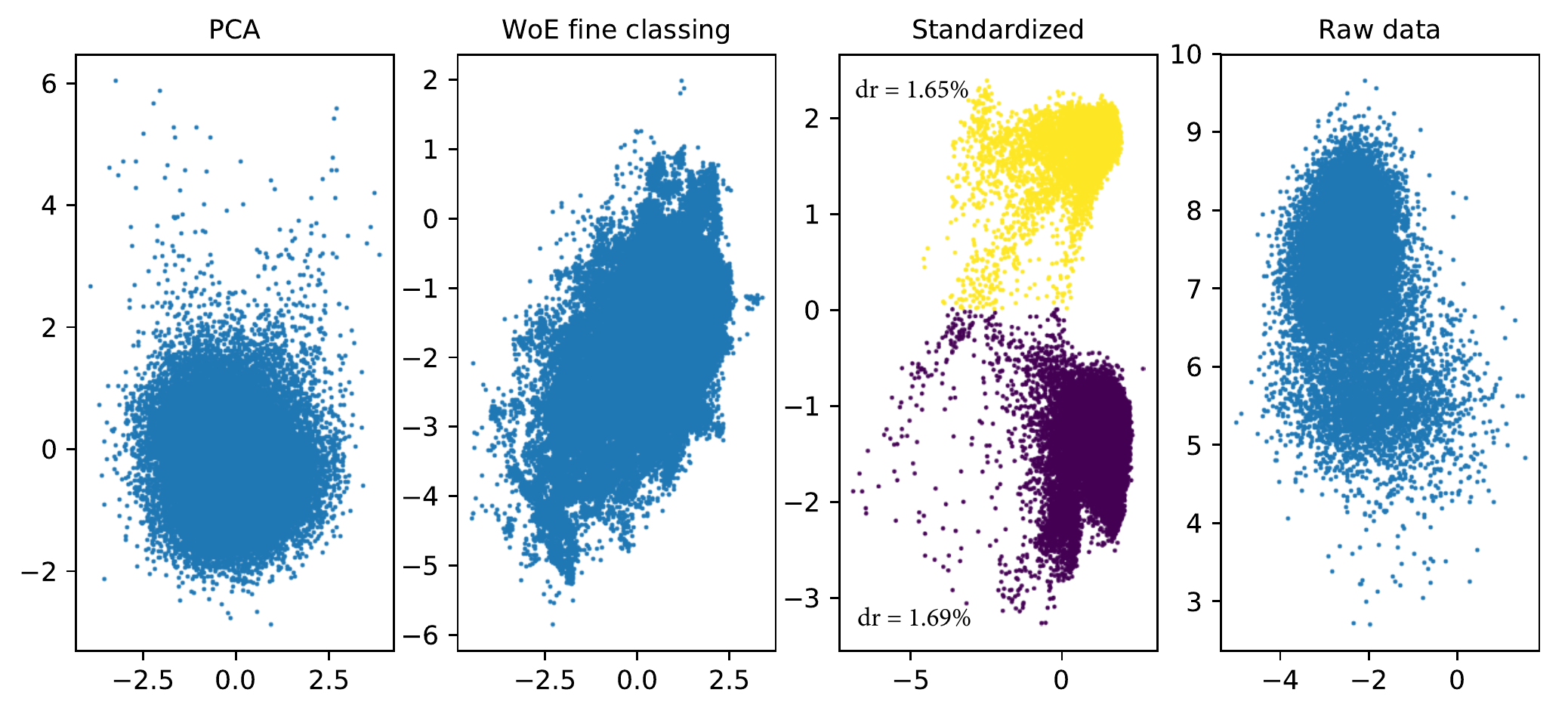}	
	\caption{Latent space for four different data transformation for the Norwegian data set. The left panel shows a PCA transformation (preserving the original data dimensionality). The second panel shows the latent space for the fine classing WoE transformation. The third panel shows the latent space for the standardized data, and finally, the right panel shows the latent space for the raw data. Standardizing the data reveals two clusters in the latent space. However, these clusters have practically the same default rate ($dr$). The other three transformations do not show any clustering structure.}
	\label{four_trans}
\end{figure}

Now we want to show that the WoE has valuable information for creating a specific grouping of the input data. Hence, we use different data transformations and, for each of these transformations, we train a new VAE, i.e. for each data transformation, we learn a data representation of the transformed input data using the same architecture in the VAE as in the one used to learn the data representations in Figure \ref{fig_latentspace}. Specifically, we generate the latent space for the following data transformations:

\begin{enumerate}
    \item PCA: The input data is transformed using principal component analysis with all the principal components, i.e. there is no dimensionality reduction.
    
    \item Standardization: The input data is standardized by removing the mean and scaling to unit variance.
    
    \item Fine classing WoE: The input data is transformed into WoE by creating bins with an approximately equal number of customers, i.e. no coarse classing is done by bank analysts.    
    
    \item Input data: Raw data without any transformation.
\end{enumerate}
Figure \ref{four_trans} shows the resulting latent spaces for the data transformations explained above. Interestingly, three of these transformations do not show any clustering structure at all. For the standardized transformation, the clusters have practically the same default rate. Hence, by identifying appealing data transformations and a useful grouping of the input data, it is possible to steer configurations in the latent space of the VAE. In this particular case, these configurations are well-defined clusters with considerably different risk profiles. 

\subsection{Cluster Interpretation}\label{sec_interp}
For the bank industry it is essential to understand which features are most important for the clustering results. To investigate this, we adopt the salient dimension methodology presented in \cite{azcarraga2005extracting} and explained in the Appendix \ref{sec_salient_dim}. This approach identifies features whose values are statistically significant in different clusters, and are called salient dimensions. In what follows, we analyse the salient dimensions for the Norwegian data set. Salient dimensions for the Kaggle and Finnish data sets can be found in Table \ref{tbl_saldim}.

The first interesting result in Figure \ref{fig_latentspace} is the pattern of the latent variables for clusters 1 and 5 (both clusters have $\text{default rate} = 5.30\%$), which are located on opposite sides of the two-dimensional space. The salient dimension \textit{MaxBucket12} in cluster 1 shows that about $70\%$ of the customers were between 30 and 60 days past due at the moment they applied for the loan, i.e. they are existing customers applying for a new loan. Actually, all customers in cluster 1 are existing customers who are at least 30 days past due. On the other hand, about $51\%$ of the customers in cluster 5 are new applicants. Cluster 2 is also composed of existing customers only. Hence, new applicants lie on the right side, while existing customers on the left hand side of Figure \ref{fig_latentspace}.

Now let us see what characterizes cluster 3, which is the cluster with the lowest default rate. Looking at the salient dimension \textit{DownPayment\%}, we can see that the average down payment in this cluster is about $20\%$, while for the rest of the clusters the average down payment is less than $12\%$. Further, the salient dimension \textit{AgeObject} shows that about $35\%$ of the customers in cluster 3 are applying to buy relatively new cars. In contrast, the average percentage of customers, in the other clusters, applying to buy new cars is about $23\%$.
 
Cluster 2 has the highest default rate and can be explained by its salient dimension \textit{MaxBucket12}. About $93\%$ of customers in this cluster are between 1 and 90 days past due, while the percentage of customers in the other clusters in the same interval is only $15\%$. 

Therefore, given that the VAE has learned a business intuitive data representation, the clusters identified for the Norwegian car loan data set can be useful for marketing campaigns, customer relationship management, data and customer management and for credit scoring \cite{anderson2007credit}. 

\section{Conclusion}\label{sec_conclusion}
In this paper, we show that it is possible to steer configurations in the latent space of the Variational Autoencoder (VAE) by transforming the input data and creating a specific grouping of it. Specifically, the Weight of Evidence (WoE) transformation encapsulates the propensity to fall into financial distress and the latent space in the VAE learns a data representation, which shows a well-defined clustering structure that encapsulates the customers' creditworthiness. 

The data representations generated with the VAE express general priors that are particularly useful in the bank industry. Specifically, the data representations are able to learn the natural clustering and spatial coherence of creditworthiness in customers data. 

Finally, our proposed method has the advantage of learning a latent data representation, which captures non-linear relationships and, for low dimensional spaces, it can be visualized. In addition, the number of clusters is suggested by the learned representation itself. Furthermore, the VAE can generate the latent configuration of new customers and assign them to one of the existing clusters. This data representation of bank customers, given that their salient dimensions are business intuitive, can be used for marketing, customer, and model fit purposes in the bank industry.

\clearpage
\section*{Acknowledgements}
The authors would like to thank Santander Consumer Bank for financial support and the real data sets used in this research. This work was also supported by the Research Council of Norway [grant number 260205] and SkatteFUNN [grant number 276428].

\section*{Appendices}
\renewcommand{\thefigure}{A\arabic{figure}}
\setcounter{figure}{0}
\renewcommand{\thetable}{A\arabic{table}}
\setcounter{table}{0}

\addcontentsline{toc}{section}{Appendices}
\renewcommand{\thesubsection}{\Alph{subsection}}

\subsection{Figures and Tables}
\begin{figure}[ht]
    \centering
    {\includegraphics[width=5cm]{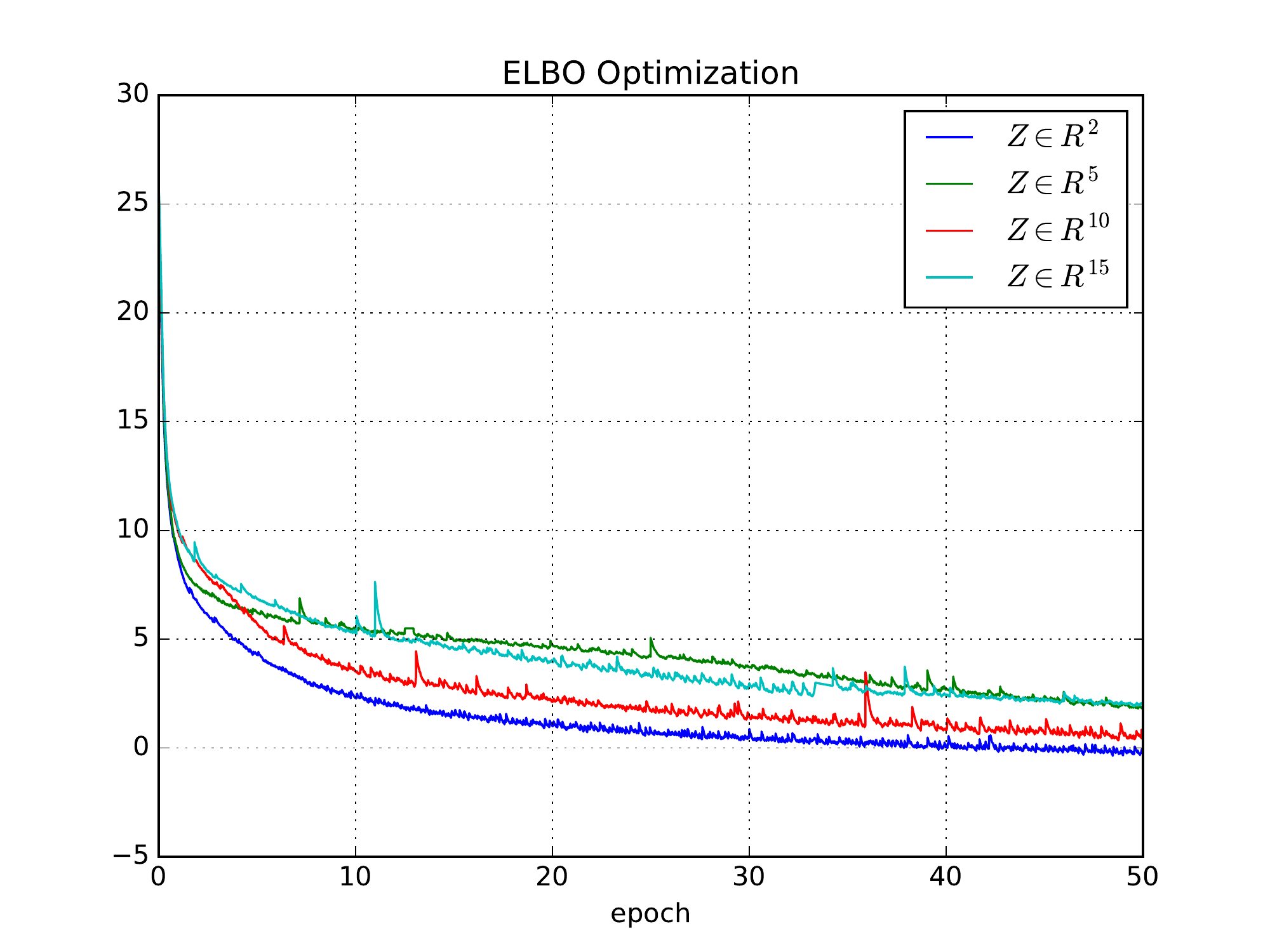}}
	\hspace*{4em}
	{\includegraphics[width=5cm]{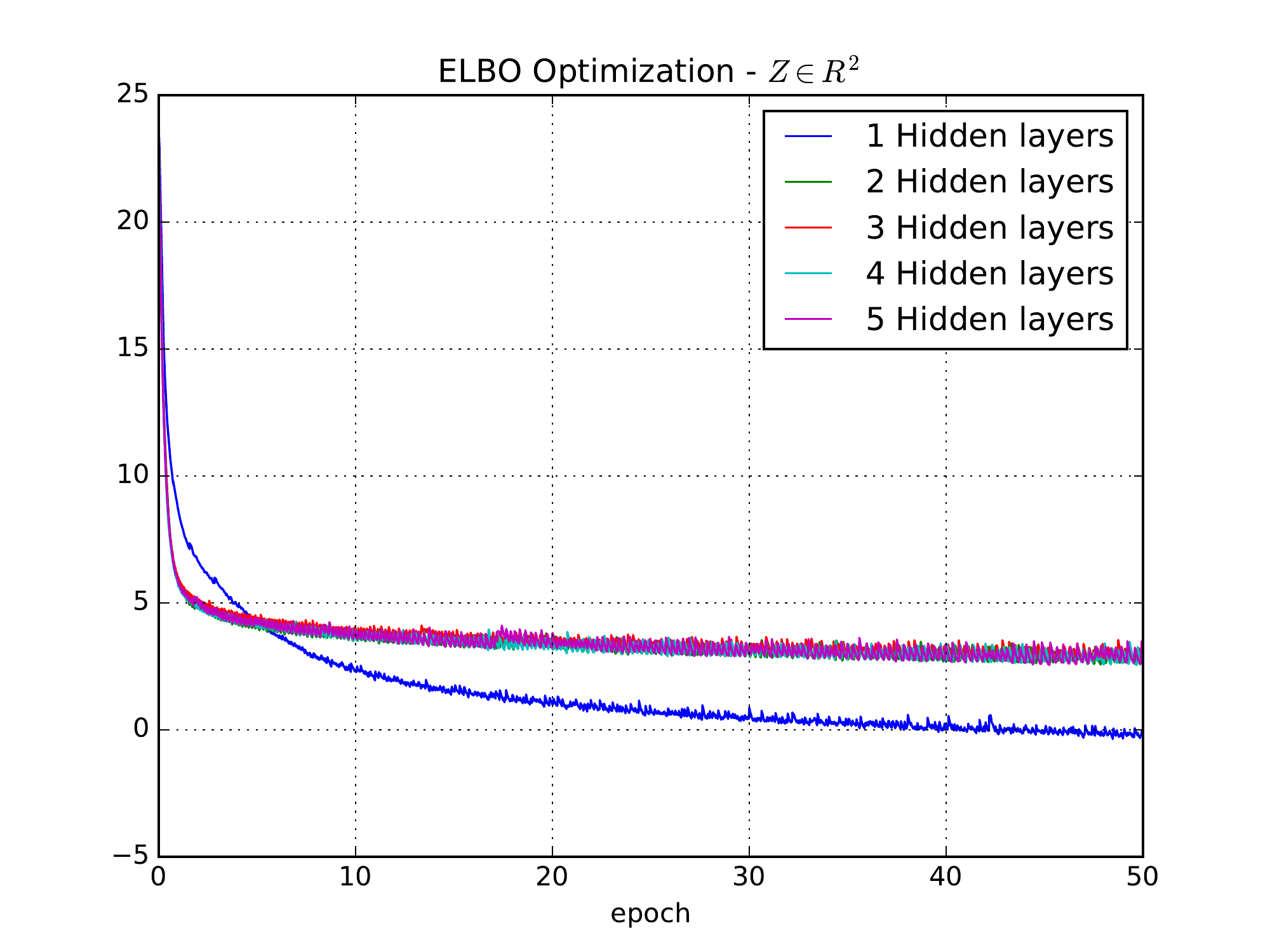}}
	\centering
	{\includegraphics[width=5cm]{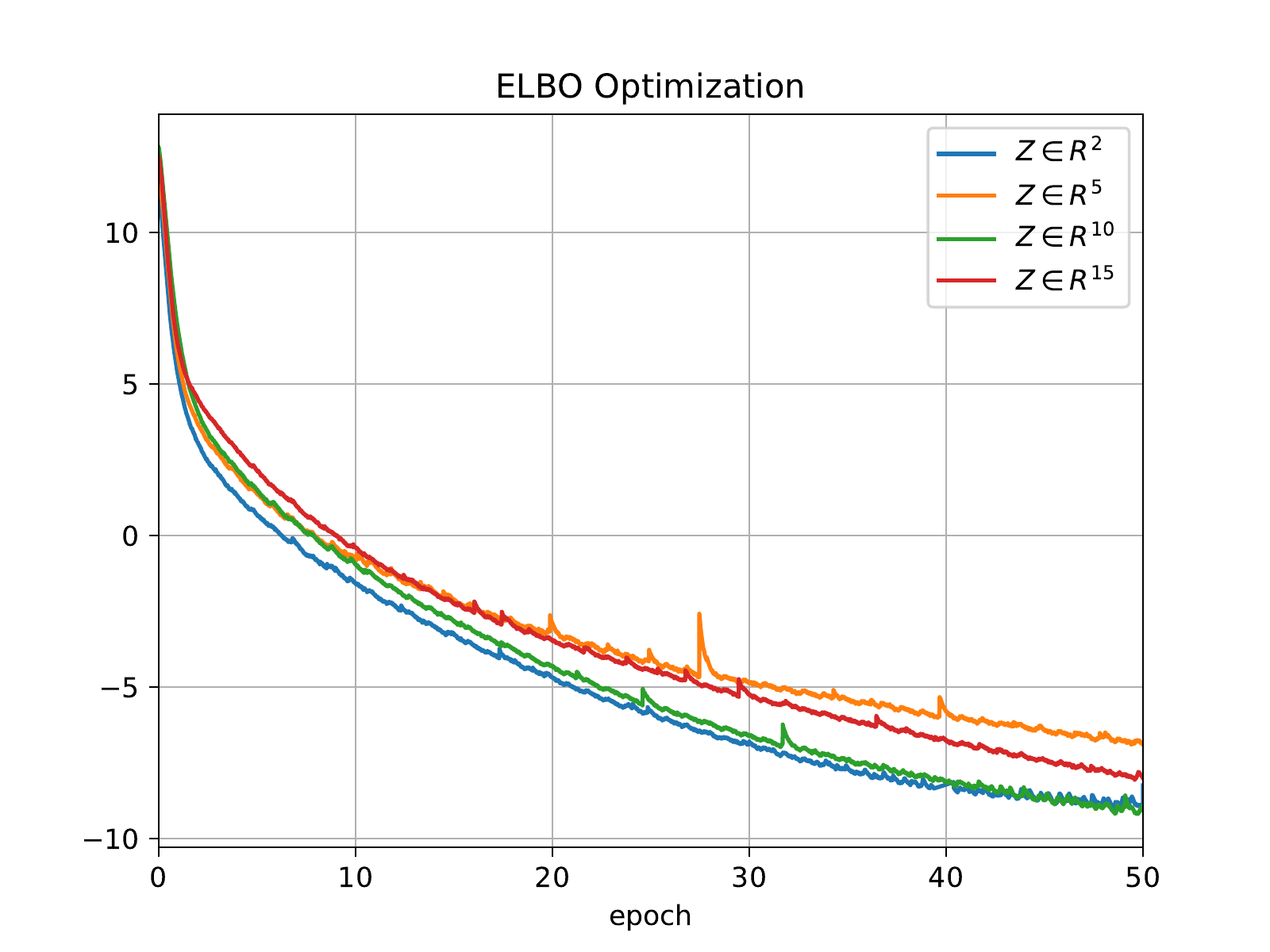}}
	\hspace*{4em}
	{\includegraphics[width=5cm]{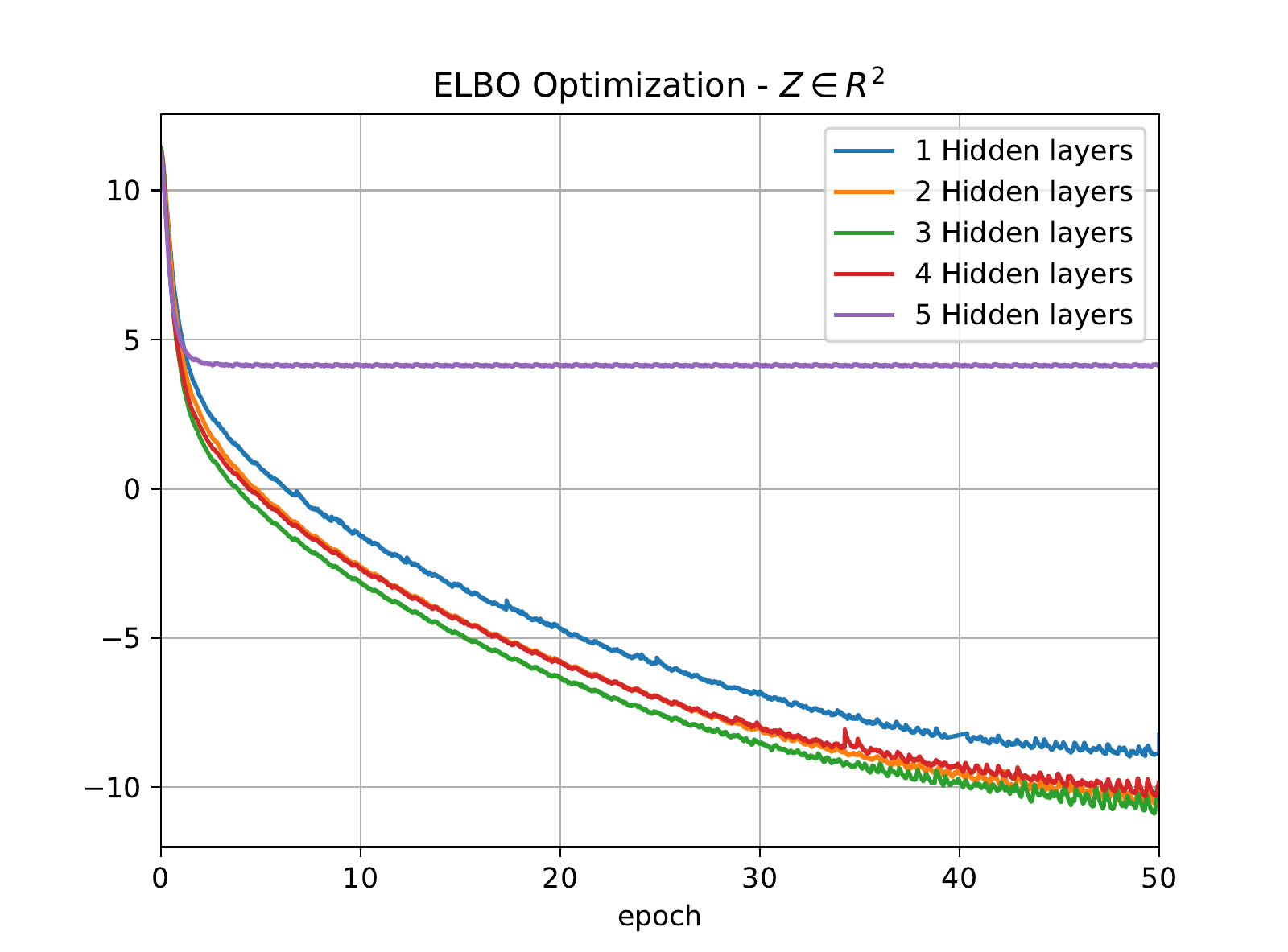}}
	\centering
	{\includegraphics[width=5cm]{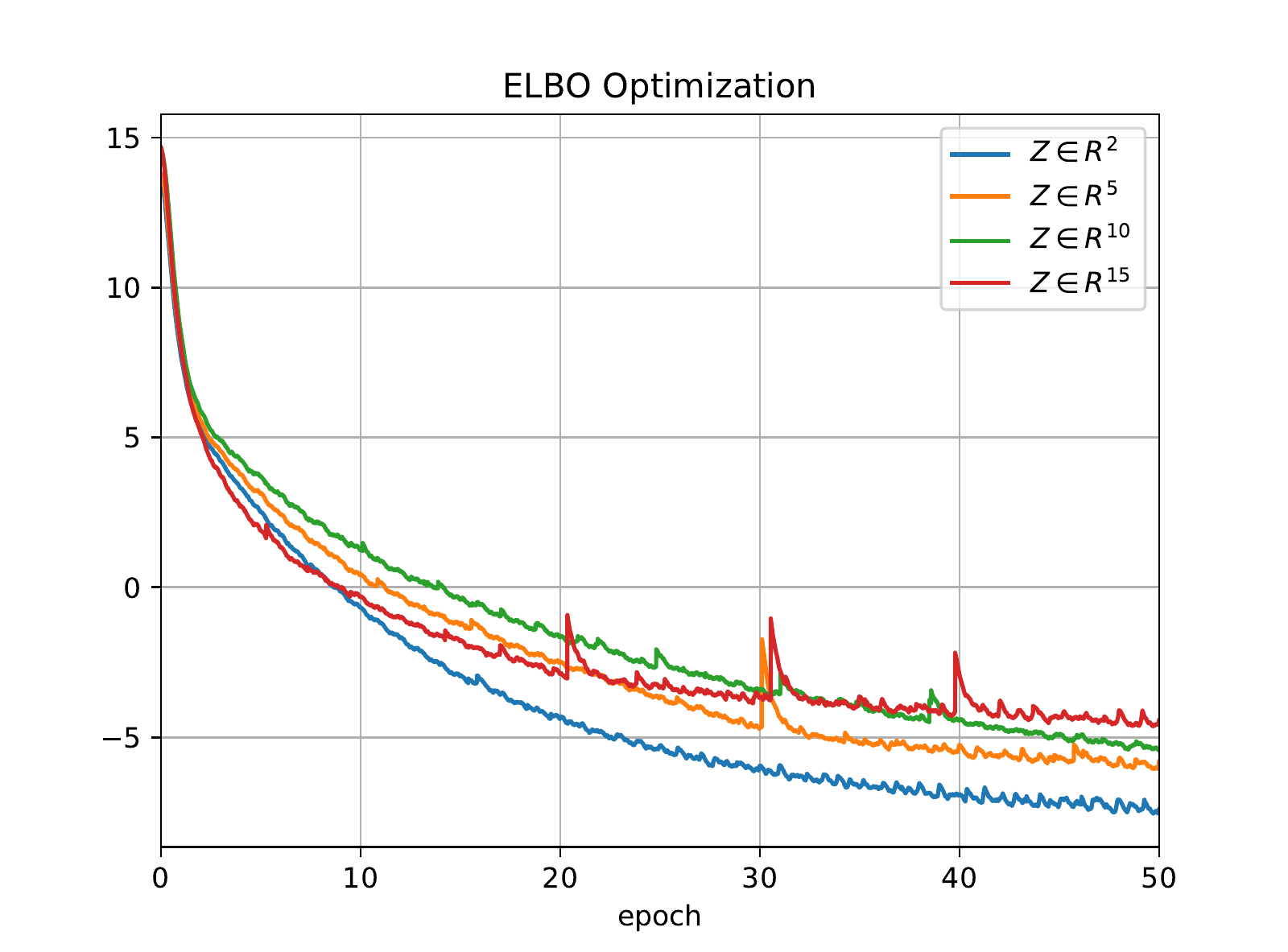}}
	\hspace*{4em}
	{\includegraphics[width=5cm]{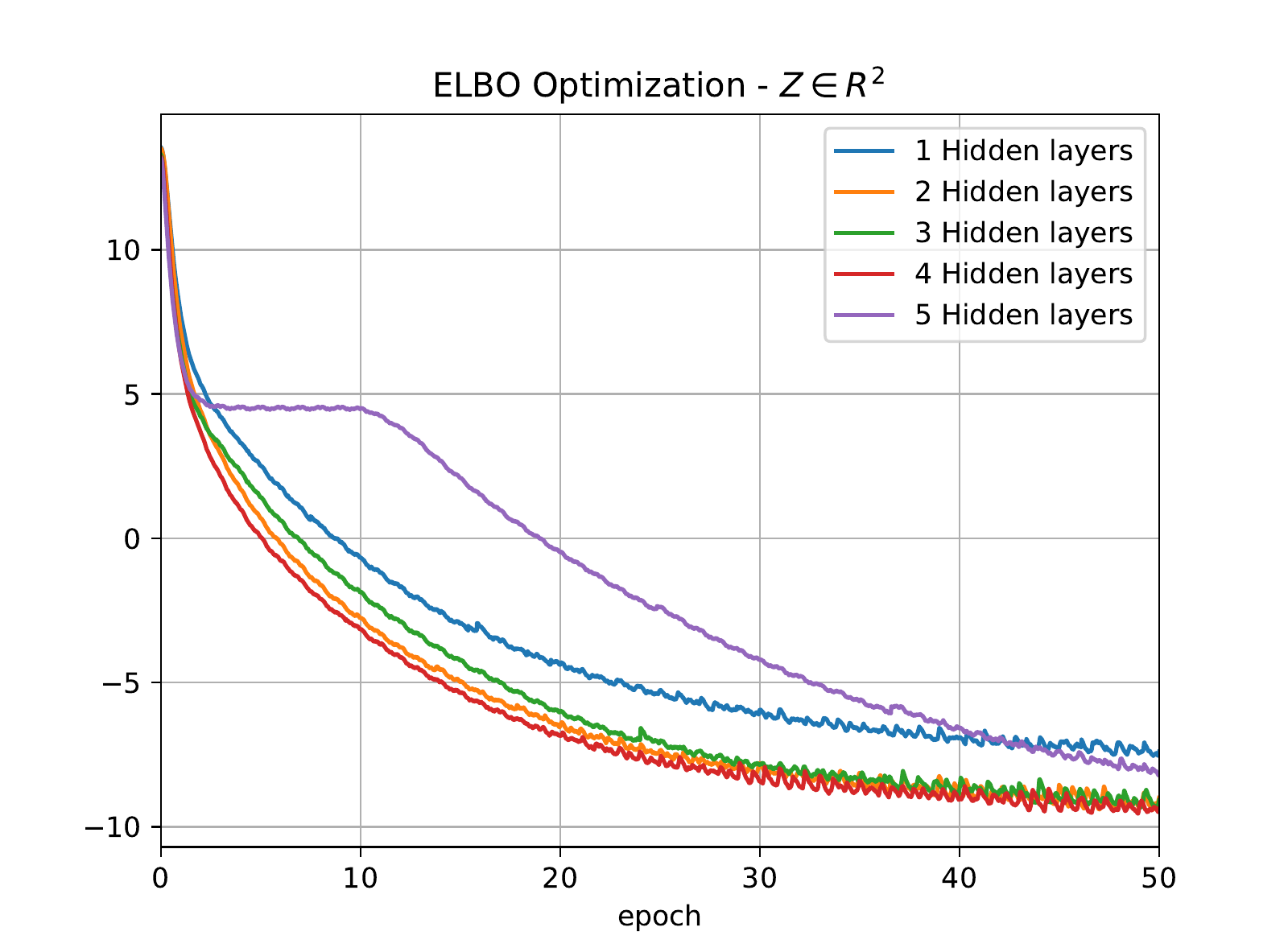}}
	\caption{Panels to the left show the optimization of the negative ELBO for different dimensionalities in the latent space. For $z \in \mathbb{R}^2$, the AEVB algorithm converges faster to the optimal variational density $q^*(z)$ for all data sets (Norwegian data set top-left panel, Kaggle data set middle-left and Finnish data set bottom-left panel). Further, panels to the right also show the optimization of the ELBO but for  $z \in \mathbb{R}^2$ and for a different number of hidden layers. The VAE for the Norwegian data set (top-right panel) with 1 hidden layer converges faster to $q^*(z)$. For the Kaggle data set (middle-right panel), 2-4 hidden layers converge faster to the optimal variational density. However, the resulting clustering structure in the latent space contains only two clusters. Similarly, for the Finnish data set (bottom-right panel) 2-4 hidden layers makes the algorithm converge faster. However, the resulting clustering structure contains four clusters. For this data set, it is not optimal to have four clusters.}
	\label{elbo_op}
\end{figure}

\begin{table}[ht!]
\centering 
\begin{adjustbox}{width=8cm}
\begin{tabular}{ |c|c|c|c|c|c| } 
\hline
Architecture ID & z dimension & Hidden Layers & Neurons & Learning Rate & Epochs \\
\hline
arch1  & 2 & 1 & 5  & 0.01  & 50 \\
arch2  & 2 & 1 & 10 & 0.01  & 50 \\
arch3  & 2 & 1 & 20 & 0.01  & 50 \\
arch4  & 2 & 1 & 30 & 0.01  & 50 \\
arch5  & 2 & 1 & 40 & 0.01  & 50 \\
arch6  & 2 & 1 & 50 & 0.01  & 50 \\
arch7  & 2 & 1 & 60 & 0.01  & 50 \\
arch8  & 2 & 1 & 70 & 0.01  & 50 \\
arch9  & 2 & 1 & 30 & 0.007 & 50 \\
arch10  & 2 & 1 & 30 & 0.008 & 50 \\
arch11 & 2 & 1 & 30 & 0.009 & 50 \\
arch12 & 2 & 1 & 30 & 0.011 & 50 \\
arch13 & 2 & 1 & 30 & 0.012 & 50 \\
arch14 & 2 & 1 & 30 & 0.013 & 50 \\
arch15  & 5 & 1 & 30 & 0.01 & 50 \\
arch16  & 10 & 1 & 30 & 0.01 & 50 \\
arch17  & 15 & 1 & 30 & 0.01 & 50 \\
arch18  & 2 & 2 & 30 & 0.01 & 50 \\
arch19  & 2 & 3 & 30 & 0.01 & 50 \\
arch20  & 2 & 4 & 30 & 0.01 & 50 \\
arch21  & 2 & 5 & 30 & 0.01 & 50 \\
\hline
\end{tabular}
\end{adjustbox}
\caption{Different architectures tested to train the VAE for the three different data sets. More complex architectures, with more hidden layers and different dimension in the latent spaces, were also tested. However, for the data sets under analysis relative complex architectures do not add any significant value.}
\label{all_archs}
\end{table}

\begin{table}[h!]
\centering 
\begin{adjustbox}{width=5.5cm}
\begin{tabular}{ |c|c|c|c| } 
\hline
Name                & Cases   & Features & Default rate \\
\hline
Norwegian car loans  &  187 069  & 20       & 0.0137     \\
Finnish car loans     &  115 899  & 12       & 0.0081  \\ 
Give me some credit  &  150 000  & 10       & 0.0668    \\
\hline
\end{tabular}
\end{adjustbox}
\caption{Summary of the three data sets used in the different experiments in this paper. Default rate for the \textit{j}'th set of customers is defined as $dr_{C_j} = \frac{\sum_i^{n_j} [y_{j,i}=1]}{n_j}$, where $n_j$ is the total number of customers and $y_i$ is the class label.}
\label{tbl_summary_dta}
\end{table}

\begin{table}[ht!]
\centering 
\begin{adjustbox}{width=7cm,totalheight=5cm}
\begin{tabular}{ |c|c| } 
\hline
\multicolumn{2}{|c|}{\textbf{\large{Norwegian car loans}}} \\
\hline
Variable Name           & Description                                  \\
\hline
BureauScoreAge          & Matrix with bureau scores and applicants age  \\
NetincomeStability      & Net income stability index                    \\
RiskBucketHistory       & Delinquency history                           \\
NumApps6M               & Number of applications last 6 months          \\
ObjectGroupCarMake      & Car brand in the application                  \\
DownPaymentAgeObject    & Matrix with down payment and car model year   \\
CarPrice                & Car price                                     \\
NetIncomet0t1           & Change in applicant's net income              \\
MaxBucketSnapshot       & Delinquency at the time of application        \\
MaxMoB12                & Months on books at the time of application    \\
NetIncomeTaxt0          & Ratio between net income and taxes            \\
AgeObject               & Car model year                                \\
AgePrimary              & Age of primary applicant                      \\
BureauScoreUnsec        & Bureau score unsecured                        \\
DownPayment             & Own capital                                   \\
MaxBucket12             & Maximum delinquency in the past 12 months     \\
TaxAmountt0             & Tax amount paid                               \\
BureauScore             & Bureau score generic                          \\
Taxt0t1                 & Change in applicant's taxes                   \\
Netincomet0             & Net income at the time of application         \\
\hline
\end{tabular}
\end{adjustbox}
\caption{Variable name and description of all features in the Norwegian car loan data set.}
\label{tbl_variables_no}
\end{table}

\begin{table}[ht!]
\centering 
\begin{adjustbox}{width=1\textwidth}
\begin{tabular}{ |c|c|c|c| } 
\hline
\multicolumn{2}{|c}{\textbf{\large{Kaggle}}}    & \multicolumn{2}{|c|}{\textbf{\large{Finnish car loans}}}            \\
\hline
Variable Name                           & Description                                   & Variable Name         & Description                                   \\
\hline
RevolvingUtilizationOfUnsecuredLines    & Total balance on credit lines                 & AgePrimary                        & Age of primary applicant                          \\
AgePrimary                              & Age of primary applicant                      & AgeObjectContractTerm             & Matrix with car model year and number of terms    \\
NumberOfTime3059DPD                     & Number of times borrower has been 30-59 dpd   & DownPayment                       & Own capital                                       \\
Monthly debt payments divided by monthly gross income  & MaritalStatus                  & Marital Status                    & DebtRatio                                         \\
Income                                  & Monthly Income                                & MaxBucket24                       & Maximum delinquency in the past 24 months         \\
NumberOfOpenCreditLines                 & Number of loans or credit cards)              & MonthsAtAddress                   & Number of months living at current address        \\
NumberOfTimesDaysLate                   & Number of times borrower has been 90 dpd      & Number2Rem                        & Number of 2nd reminders last year                 \\
NumberRealEstateLoansOrLines            & Number of mortgage loans                      & NumberRejectedApps                & Number of rejected applications                   \\
NumberOfTime6089DPD                     & Number of times borrower has been 60-89 dpd   & ObjectPrice                       & Car price                                         \\
NumberOfDependents                      & Number of dependents in family                & ResidentialStatus                 & Whether the applicant owns a house                \\
                                        &                                               & ObjectMakeUsedNew                 & Matrix with car make and whether it is new or used\\
                                        &                                               & EquityRatio                       & Debt to equitity                                  \\
\hline
\end{tabular}
\end{adjustbox}
\caption{Variable name and description of all features in the Kaggle and Finnish car loan data sets.}
\label{tbl_variables}
\end{table}

\clearpage

\begin{table}[ht!]
\centering 
\begin{adjustbox}{width=10cm,totalheight=4cm}
\begin{tabular}{ |c|c|c|c|c|c| } 
\hline
\multicolumn{2}{|c|}{\textbf{\large{Norwegian car loan}}} & \multicolumn{2}{c|}{\textbf{\large{Kaggle}}} & \multicolumn{2}{c|}{\textbf{\large{Finnish car loan}}} \\
\hline
Cluster  & Salient Dimension & Cluster  & Salient Dimension & Cluster  & Salient Dimension \\
\hline
1 &  MaxBucket12    & 1 & NumberOfTime3059DPD                   & 1 & AgePrimary        \\
2 &  NetIncomet0t1  & 1 & NumberOfTimesDaysLate                 & 1 &  Number2Rem       \\ 
2 &  MaxBucket12    & 1 & NumberRealEstateLoansOrLines          & 1 & NumberRejectedApps\\
3 &  AgeObject      & 1 & NumberOfTime6089DPD                   & 2 & Number2Rem        \\ 
3 &  NetIncomet0t1  & 2 & RevolvingUtilizationOfUnsecuredLines  & 2 & NumberRejectedApps\\
3 &  Taxt0t1        & 2 & DebtRatio                             & 3 & DownPayment       \\ 
3 &  DownPayment    & 3 & NumberOfTime3059DPD                   & 3 & ResidentialStatus \\ 
4 &  NumApps6M      & 3 & NumberOfTime6089DPD                   &   &                   \\
4 &  AgeObject      & 3 & NumberOfDependents                    &   &                   \\
4 &  NetIncomet0t1  & 4 & NumberOfTime3059DPD                   &   &                   \\
4 &  Taxt0t1        & 4 & NumberOfTimesDaysLate                 &   &                   \\
4 &  DownPayment    & 4 & NumberOfTime6089DPD                   &   &                   \\
5 &  AgeObject      &   &                                       &   &                   \\
5 &  NetIncomet0t1  &   &                                       &   &                   \\
5 &  DownPayment    &   &                                       &   &                   \\
\hline
\end{tabular}
\end{adjustbox}
\caption{Statistically significant salient dimensions for the Norwegian, Kaggle and Finnish data set. We use $s.d.=1$ to define salient dimensions.}
\label{tbl_saldim}
\end{table}

\subsection{Salient Dimensions}\label{sec_salient_dim}
Let $v$ be the \textit{v'th} dimension of the \textit{i'th} vector $x_{i,v}$, where $x \in R^{\ell}$. Further let $\Phi_{in}(k)$ be the set of in-patterns (within cluster $k$) and $\Phi_{out}(k)$ be the set of out-patterns (not within cluster $k$). Then compute the mean input values  
\begin{align}
    \mu_{in}(k,v) &= \frac{\sum_{x_i \in \Phi_{in}(k)} x_{i,v} }{|\Phi_{in}(k)|}, \\
    \mu_{out}(k,v) &= \frac{\sum_{x_i \in \Phi_{out}(k)} x_{i,v} }{|\Phi_{out}(k)|},
\end{align}
where $|\{\cdot\}|$ returns the cardinality of $\{\cdot\}$. Further, compute the difference factors 
\begin{equation}
    df(k,v)=\frac{\mu_{in}(k,v)-\mu_{out}(k,v)}{\mu_{out}(k,v)},
\end{equation}
and their mean and standard deviations
\begin{align}
    \mu_{df}(k) &= \frac{1}{\ell}\sum_v^\ell df(k,v),  \\
    \sigma_{df}(k) &= \sqrt{\sum_v^\ell\big(df(k,v)-\mu_{df}(k)\big)^2/\ell}.
\end{align}
Finally, we say that the \textit{v'th} feature in cluster $k$ is a salient dimension if
\begin{equation}
    df(k,v) \leq \mu_{df}(k) - s.d. \ \sigma_{df}(k),
\end{equation}
or
\begin{equation}
    df(k,v) \geq \mu_{df}(k) + s.d. \ \sigma_{df}(k), 
\end{equation}
where $s.d.$ is the number of standard deviations to be used. The value for $s.d$ is defined based on the data set. We use $s.d.=1$ for all three data sets under analysis.

\end{document}